\documentclass[sigconf]{acmart}
\usepackage{multirow}
\usepackage[normalem]{ulem}
\useunder{\uline}{\ul}{}
\usepackage{subfigure}
\usepackage{colortbl}
\newcommand{\code}[0]{\url{https://github.com/ZzoomD/FairINV/}}
\usepackage{algorithm}
\usepackage{algorithmic}
\usepackage{balance}

\AtBeginDocument{%
  \providecommand\BibTeX{{%
    \normalfont B\kern-0.5em{\scshape i\kern-0.25em b}\kern-0.8em\TeX}}}

\copyrightyear{2024}
\acmYear{2024}
\setcopyright{acmlicensed}\acmConference[KDD '24]{Proceedings of the 30th ACM SIGKDD Conference on Knowledge Discovery and Data Mining}{August 25--29, 2024}{Barcelona, Spain}
\acmBooktitle{Proceedings of the 30th ACM SIGKDD Conference on Knowledge Discovery and Data Mining (KDD '24), August 25--29, 2024, Barcelona, Spain}
\acmDOI{10.1145/3637528.3672029}
\acmISBN{979-8-4007-0490-1/24/08}

\begin{document}

\title{One Fits All: Learning Fair Graph Neural Networks \\for Various Sensitive Attributes}

\author{Yuchang Zhu}
\affiliation{
  \institution{Sun Yat-sen University}
  \city{Guangzhou}
  \country{China}
}
\email{zhuych27@mail2.sysu.edu.cn}

\author{Jintang Li}
\affiliation{%
  \institution{Sun Yat-sen University}
  \city{Guangzhou}
  \country{China}
  }
\email{lijt55@mail2.sysu.edu.cn}

\author{Yatao Bian}
\affiliation{%
  \institution{Tencent AI Lab}
  \city{Shenzhen}
  \country{China}
  }
\email{yatao.bian@gmail.com}

\author{Zibin Zheng}
\affiliation{%
 \institution{Sun Yat-sen University} 
 \city{Guangzhou}
 \country{China}
 }
\email{zhzibin@mail.sysu.edu.cn}

\author{Liang Chen}
\authornote{Corresponding author.}
\affiliation{%
  \institution{Sun Yat-sen University}
  \city{Guangzhou}
  \country{China}
}
\email{chenliang6@mail.sysu.edu.cn}

\renewcommand{\shortauthors}{Yuchang Zhu, Jintang Li, Yatao Bian, Zibin Zheng, \& Liang Chen}

\begin{abstract}
  Recent studies have highlighted fairness issues in Graph Neural Networks (GNNs), where they produce discriminatory predictions against specific protected groups categorized by sensitive attributes such as race and age. While various efforts to enhance GNN fairness have made significant progress, these approaches are often tailored to specific sensitive attributes. Consequently, they necessitate retraining the model from scratch to accommodate changes in the sensitive attribute requirement, resulting in high computational costs. To gain deeper insights into this issue, we approach the graph fairness problem from a causal modeling perspective, where we identify the confounding effect induced by the sensitive attribute as the underlying reason. Motivated by this observation, we formulate the fairness problem in graphs from an invariant learning perspective, which aims to learn invariant representations across environments. Accordingly, we propose a graph fairness framework based on invariant learning, namely \textbf{FairINV}, which enables the training of fair GNNs to accommodate various sensitive attributes within a single training session. Specifically, FairINV incorporates sensitive attribute partition and trains fair GNNs by eliminating spurious correlations between the label and various sensitive attributes. Experimental results on several real-world datasets demonstrate that FairINV significantly outperforms state-of-the-art fairness approaches, underscoring its effectiveness. \footnote{Our code is available via: \code.}
\end{abstract}




\maketitle

\section{Introduction}
% background of GNN, GNN's fairness problem.
Graph neural networks (GNNs) have achieved tremendous success in processing graph-structured data~\cite{kipf2016semi,maskgae,spikegcl}, such as citation networks~\cite{shchur2018pitfalls} and social networks~\cite{leskovec2012learning,takac2012data}. Consequently, this advancement has led to their application across diverse domains, including fraud detection~\cite{dou2020enhancing} and recommender systems~\cite{wu2022graph}. However, recent studies~\cite{dai2021say,spinelli2021fairdrop} have unveiled a concerning trend that GNNs make discriminatory decisions toward the specific protected groups defined by sensitive attributes, e.g., race, and age. This phenomenon, termed the group fairness problem of GNNs, hinders the application of GNNs in high-stake scenarios.

\begin{figure*}[!ht]
  \centering
  \includegraphics[width=0.9\textwidth]{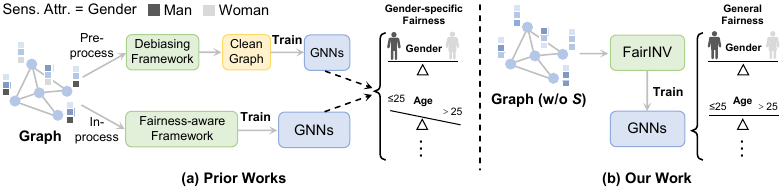}
  \caption{An illustration of comparison between prior works and our work (FairINV). (a) Prior works (pre-process and in-process) achieve fairness toward the specific sensitive attribute; (b) Our work trains fair GNNs toward various sensitive attributes in a single training session without accessing the sensitive attribute.}
  % b.不需要敏感属性且面向多个敏感属性；c.公平性问题是通过两个渠道影响的，现有方法只能消除一个渠道的影响，而我们的方法可以消除两个渠道的影响
  \label{fig:intro}
\end{figure*}

% the existing method and its problem: tailored for a specific sensitive attribute, With a toy example. 
To improve the fairness of GNNs, considerable efforts have been devoted to debiasing the training data~\cite{dong2022edits,ling2022learning} or learning fair GNNs directly~\cite{dai2021say,bose2019compositional}, referred to as the pre-process and in-process approaches, respectively. Within these two methodological categories, common implementations encompass adversarial learning~\cite{ling2022learning,wang2022improving}, distribution alignment among various protected groups~\cite{dong2022edits,guo2022learning}, graph-structured data modification~\cite{spinelli2021fairdrop,ling2022learning,dong2022edits}, and edge reweighting~\cite{li2021dyadic}. Despite significant progress, these approaches are often tailored to specific sensitive attributes, as shown in Figure~\ref{fig:intro}(a). Consequently, training GNN models from scratch becomes imperative when faced with fairness requirement alterations in sensitive attributes, such as transitioning from age-based considerations to gender-related factors. Take loan approvals in a credit card network as an example, according to fairness policies, initially trained GNNs are designed to make fair decisions toward the protected groups divided by age, e.g., age $\leq$ 25 and age $\textgreater$ 25. However, when policies change to focus on gender, necessitating fair treatment between male and female groups, the previously tailored model optimized for age fairness becomes inadequate. Hence, this mandates retraining the GNN model to ensure fairness regarding gender, which is a laborious and computationally intensive process.  

% the core idea of this work.

% to solve the above problem, the requirements are as follows. 
In summary, there is a significant demand for a universal graph fairness approach that trains fair GNNs across various sensitive attributes in a single training session. Achieving such an approach entails addressing the following challenges: (1) \textbf{Generalization to various sensitive attributes.} Previous studies aim to achieve fairness tailored for the specific sensitive attribute. Additionally, these approaches always require accessing the sensitive attributes in the training process, which is impractical in real-world scenarios due to legal limitations~\cite{lahoti2020fairness,chai2022fairness}. Correspondingly, our first challenge is to design a fairness framework that achieves fairness w.r.t. various sensitive attributes in a single training session without accessing the sensitive attribute, as shown in Figure~\ref{fig:intro}(b). (2) \textbf{Full fairness.} %Inspired by the fairness study in face recognition~\cite{ma2023invariant}, we present a causal graph in Figure~\ref{fig:intro}(c) to further understand the fairness problem on graph-structured data.
According to Section~\ref{subsec:cas}, two causal pathways ($S$ $\rightarrow$ $Y$ and $S \rightarrow \mathcal{G} \rightarrow Y$) demonstrate how the sensitive attribute $S$ influences the labels $Y$, misleading the trained GNNs to capture the sensitive attribute information for predictions. In this regard, $S$ is a confounder. 
%As such, the trained GNNs capture spurious correlations between $S$ and $Y$, resulting in discriminatory predictions against the sensitive attribute. 
To achieve full fairness, blocking these two causal effects appears to be a straightforward solution. However, it is challenging due to the presence of underlying spurious correlations between unobservable variables $S$ and $Y$. As discussed in Section~\ref{subsec:cas}, prior works failed to eliminate both causal effects concurrently. Inspired by INV-REG~\cite{ma2023invariant}, backdoor adjustment implemented by data partition presents a promising approach to tackle this challenge.

% contributions
In this work, we first formulate the graph fairness issue from an invariant learning~\cite{wu2022handling,chen2022learning,li2022learning} perspective, where sensitive attributes as environments. Building upon this formulation, we propose a universal graph fairness framework named \textbf{FairINV}. To overcome the first challenge, FairINV jointly optimizes a fair GNN for multiple sensitive attributes inferred automatically via sensitive attribute partition. To overcome the second challenge, FairINV incorporates invariant learning optimization objectives building upon sensitive attribute partition to remove confounding effects induced by $S$. Specifically, the optimization objective of FairINV gives rise to equal predictions of trained GNNs across environments (sensitive attributes). In summary, FairINV mitigates spurious correlations between various sensitive attributes and the label. Our contributions can be summarized as follows:

\begin{itemize}
    \item We study the fairness issue on graphs from an invariant learning perspective. To the best of our knowledge, this is the first attempt to explore graph fairness from this particular perspective.
    \item We introduce FairINV, a universal graph fairness framework that inherits the spirit of graph invariant learning. An unsupervised sensitive attributes partition of FairINV facilitates fairness improvement in terms of various sensitive attributes. 
    \item We conduct experiments on several real-world datasets to validate the effectiveness of FairINV. Experimental results show that FairINV can train a fair GNN toward various sensitive attributes in a single training session.
\end{itemize}

\section{Related Work}

\subsection{Fairness in Graph Neural Networks}
% 注意要区分ICML19 CFC这篇论文
%GNNs have emerged as a powerful technology and their fairness has become a rising concern. 
The fairness of GNNs includes group fairness~\cite{dai2021say,agarwal2021towards,zhu2024fair} and individual fairness~\cite{kang2020inform,dong2021individual}. Our study focuses on the group fairness aspect, emphasizing equitable model decisions for each protected group partitioned by the sensitive attribute. %Individual fairness, on the other hand, underscores similar treatment for similar individuals. Group fairness represents the most extensively studied fairness notion and is the focus of investigation in this work. 
Recent studies improving group fairness in GNNs typically segregate into pre-process~\cite{dong2022edits,ling2022learning,zhu2024fairagg} and in-process~\cite{wang2022improving,bose2019compositional,zhu2024devil} approaches. Pre-process approaches aim to mitigate biases in training data before training downstream tasks. To mitigate biases, techniques like adversarial learning~\cite{ling2022learning}, and distribution alignment~\cite{dong2022edits,current2022fairmod,yang2022obtaining} serve as optimization objectives for debiasing training data. Additionally, some heuristic approaches modify the training graph~\cite{spinelli2021fairdrop,li2022fairlp} or reweight edge~\cite{khajehnejad2022crosswalk,li2021dyadic} by either enhancing connections between diverse groups or reducing connections within the same groups. In-process approaches aim to train fair GNNs through the fairness-aware framework. Similar to pre-process approaches, in-process approaches also incorporate adversarial learning~\cite{bose2019compositional,dai2021say} and distribution alignment~\cite{guo2022learning,fan2021fair} to learn GNNs. Despite significant progress, these approaches are tailored to the specific sensitive attribute, lacking considerations for various sensitive attributes. 

Despite Bose et al.'s~\cite{bose2019compositional} work of a compositional adversarial framework using a set of sensitive-invariant filters, it necessitates prior knowledge of considered sensitive attributes and their specific values for each individual. In contrast, our work learns fair GNNs toward various sensitive attributes in a single training session without accessing sensitive attributes, which remains under-explored for prior works.

\subsection{Invariant Learning for Fairness}
Guided by the independent causal mechanism assumption~\cite{peters2017elements,peters2016causal}, invariant learning, capable of capturing invariances across various environments, stands as a significant approach facilitating out-of-distribution (OOD) generalization~\cite{creager2021environment,ahuja2020invariant,arjovsky2019invariant,tan2023provably}. The core idea behind invariant learning is to learn causal information that stays invariant across different environments while disregarding spurious correlations that exhibit variability~\cite{chen2022does}. However, there is limited research exploring the application of invariance learning in fairness. Adragna et al.~\cite{adragna2020fairness} empirically illustrate how invariant risk minimization in invariant learning can contribute to building fair machine learning models. Ma et al.~\cite{ma2023invariant} point out the fairness-related bias in face recognition stemming from confounding demographic attributes. Then, they iteratively partition data to annotate confounders and learn invariant features to remove the confounding effect. Yet, these explorations of invariant learning in fairness predominantly focus on Euclidean data. Conversely, significant efforts~\cite{chen2022learning,liu2023flood,li2022learning} have addressed the out-of-distribution problem within graph structures from an invariant learning perspective. However, the effectiveness of invariant learning in ensuring graph fairness remains an under-explored area. To the best of our knowledge, our work is the first to explore the graph fairness problem utilizing graph invariant learning. 

\section{Preliminaries}
In this section, we first introduce the detailed notations used in this work. Then, we give a causal analysis for our study problem, followed by the problem formulation of this work. 

\subsection{Notations}
 Let $\mathcal{G}=(\mathcal{V}, \mathcal{E}, \textbf{X})$ denote an undirected and unweighted attributed graph, where $\mathcal{V}$ is a set of nodes and $\mathcal{E}$ is a set of edges. Meanwhile, $\lvert \mathcal{V} \rvert = n$ and $\lvert \mathcal{E} \rvert = m$ represent the number of nodes and edges, respectively. $\textbf{X} \in \mathbb{R}^{n\times d}$ represents the node attribute matrix without the sensitive attribute $S$ where $d$ is the node attribute dimension. $\textbf{A} \in \lbrace0, 1\rbrace^{n\times n}$ is the adjacency matrix where $\textbf{A}_{uv}=1$ indicates the edge connection $e_{uv} \in \mathcal{E}$ between the node $u$ and the node $v$, and $\textbf{A}_{uv}=0$ otherwise. Nodes with the same sensitive attribute value belong to the same protected group. Most GNNs follow the message-passing mechanism, which aggregates messages from their neighbors, and can be summarized as follows:
\begin{equation}
\label{eq:gnn}
% \begin{split}
\small
  % a^{(l)}_v &= \operatorname{AGGREGATE}^{(l)}(\{h^{(l-1)}_u:u\in \mathcal{N}(v)\}),\\
    \textbf{h}^{(l)}_u = \operatorname{UPD}^{(l)}(\{\textbf{h}^{(l-1)}_u, \operatorname{AGG}^{(l)}(\{\textbf{h}^{(l-1)}_v:v\in \mathcal{N}(u)\})\}),
% \end{split}
\end{equation}
where $l$ is the layer number, $\operatorname{AGG}^{(l)}(\cdot)$ and $\operatorname{UPD}^{(l)}(\cdot)$ denote aggregation function and update function in $l$-th layer, respectively. $\mathcal{N}(u)$ denote the set of nodes adjacent to node $u$.

While our approach is applicable to various downstream tasks, in this paper, we exemplify its application using the node classification downstream task to illustrate the proposed methodology. A GNN model $f$, consisting of an encoder $f_{g}$ and a linear classifier $f_{c}$, takes a graph $\mathcal{G}$ as input and outputs the node predicted label $\hat{Y}=f_{c}(f_{g}(\mathcal{G}))$. The goal of $f$ is to predict $\hat{Y}$ such that it is as close as possible to the ground truth labels $Y$.

\begin{figure}[!t]
  \centering
  \includegraphics[width=\linewidth]{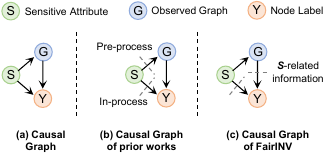}
  \caption{Structural causal model for GNNs prediction. (a) The fairness issue on graphs can be caused by two causal pathways, i.e., $S \rightarrow Y$ and $S \rightarrow \mathcal{G} \rightarrow Y$; (b) Prior works either exclusively eliminate the causal pathway $S \rightarrow \mathcal{G}$ or exclusively eradicate the causal pathway $S \rightarrow Y$; (c) FairINV tackles the fairness issue through blocking both two causal pathways.}
  % b.不需要敏感属性且面向多个敏感属性；c.公平性问题是通过两个渠道影响的，现有方法只能消除一个渠道的影响，而我们的方法可以消除两个渠道的影响
  \label{fig:causal_analysis}
\end{figure}

\subsection{Causal Analysis}
\label{subsec:cas}
To broaden insights, we construct a structural causal model~\cite{peters2017elements} (SCM) to analyze the group fairness issue in graph-structured data. Figure~\ref{fig:causal_analysis}(a) illustrates the causal relationship among the graph $\mathcal{G}$, the sensitive attribute $S$, and the node label $Y$. In the SCM, there are two causal pathways by which the sensitive attribute $S$ affects the node label $Y$, leading to issues of group fairness in GNNs. A detailed description of these two causal pathways is provided below.
\begin{itemize}
    \item $S \rightarrow \mathcal{G} \rightarrow Y$. This causal pathway describes the influence of $S$ on the formation of graph-structured data $\mathcal{G}$, which subsequently impacts the predictions of node labels $Y$ by the trained GNN. Specifically, the path $S \rightarrow \mathcal{G}$ represents the impact of $S$ on the generation process of graph-structured data. In this context, the data exhibit two primary phenomena: (1) the graph topology exhibits sensitive homophily~\cite{jiang2023chasing}, where connected nodes are more likely to share the same sensitive attribute $S$. (2) Non-sensitive node attributes may implicitly convey information about $S$. For instance, if $S$ represents gender, certain attributes like height, while not directly sensitive, become relevant in inferring an individual's gender. The path $\mathcal{G} \rightarrow Y$ encapsulates the training process of GNNs, wherein the network may inherit and subsequently propagate biases (information related to $S$) present in the training data. 
    \item $S \rightarrow Y$. This causal pathway illustrates the underlying correlation between the node label and the sensitive attribute. This correlation often originates from societal discrimination against protected groups, leading to biased predictions in the trained GNN. For instance, given a social network dataset, the task is to predict the user's occupational field. The dataset predominantly comprises occupations of females as nurses and males as engineers. Consequently, the GNN trained on this dataset tends to predict engineering as the occupational field for males and nursing for females, thus revealing a gender bias in its predictions. This phenomenon can be attributed to the $S \rightarrow Y$ causal pathway inherent in the dataset.
\end{itemize}

In summary, discriminatory decisions in GNNs stem from the two causal pathways discussed above. In this regard, the pathway $Y \leftarrow S \rightarrow \mathcal{G}$ is a backdoor path, with $S$ acting as a confounder. This pathway may mislead the trained GNN to utilize the sensitive attribute for predictions, known as the spurious confounding effect. To remove this effect, a straightforward yet challenging solution involves eliminating $S \rightarrow \mathcal{G} \rightarrow Y$ and $S \rightarrow Y$. However, prior works have not successfully removed both pathways simultaneously. As shown in Figure~\ref{fig:causal_analysis}(b), pre-process methods primarily focus on reducing the information related to the sensitive attribute in the training data, effectively removing the path $S \rightarrow \mathcal{G}$. Conversely, in-process methods strive to develop a fair GNN that makes decisions independently of $S$, akin to removing the path $S \rightarrow Y$. Another approach to mitigate confounding effects is the backdoor adjustment, achieved by partitioning the training data into different splits. In our scenarios, we partition nodes into distinct demographic groups and learn GNNs invariant across these groups. Drawing inspiration from a fairness study in face recognition~\cite{ma2023invariant}, we attempt to formulate the graph fairness issue from an invariant learning perspective. Leveraging the environment inference capabilities of invariant learning, we can unsupervisedly infer the sensitive attribute of nodes, facilitating group partitioning. 

\begin{figure*}[!ht]
  \centering
  \includegraphics[width=0.95\linewidth]{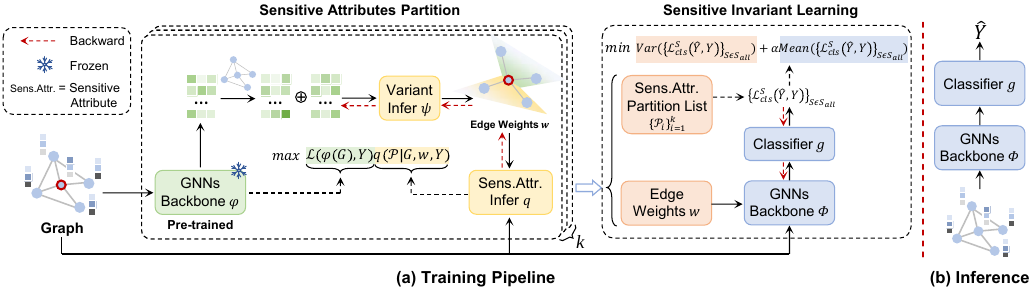}
  \caption{The overview of FairINV. FairINV includes two stages: Sensitive Attributes Partition (SAP) and Sensitive Invariant Learning (SIL).}
  \label{fig:overview}
\end{figure*}

\subsection{Problem Formulation}
\label{subsec:problem_formulate}
% Given a set of graphs $\mathcal{D}=\{\mathcal{D}^{e}\}_{e}$ collected from various environments $\mathcal{E}_{all}$, a GNN $\rho \circ h$ consisting of an encoder $h$ and a classifier $\rho$ takes these graphs as input to predict graph labels. Here, $\mathcal{D}^{e}=\{\mathcal{G}_{i}^{e},y_{i}^{e}\}$ is graphs from environment $e$. The OOD generalization aims to learn GNNs to generalize to all unseen environments. Formally, the OOD generalization on the graph level is to minimize:
% \begin{equation}
% \label{eq:ood_graph}
%   \mathop{\min}_{\rho, h}\mathop{\max}_{e \in \mathcal{E}_{all}} \mathcal{R}^{e}(\rho \circ h)
% \end{equation}
% where $\mathcal{R}^{e}$ is the empirical risk under environment $e$.
In this subsection, we formulate the graph fairness problem from an invariant learning perspective. Our work focuses on the node-level task. Following the setting of EERM~\cite{wu2022handling}, we investigate the impact of the node's ego-graph on the centered node. Specifically, given a single graph $\mathcal{G}=(\mathcal{V}, \mathcal{E}, \textbf{X})$, we have a set of ego-graph $\mathcal{D}=\mathcal{D}^{e}$ from various environment $\mathcal{E}_{all}$, where $\mathcal{D}^{e}=\{\mathcal{G}_{v}^{e},y_{v}^{e}\}$ is graphs from environment $e$. $\mathcal{G}_{v}^{e}$ and $y_{v}^{e}$ are the ego-graph and the node label of node $v$. The invariant learning aims to learn GNNs to generalize to all unseen environments. Denote $f$ as a GNN model consisting of an encoder and a classifier, $\hat{y}_{v}=f(\mathcal{G}_{v})$ as the predicted label of node $v$, and $\mathcal{R}^{e}(\cdot)$ as the empirical risk under environment $e$. Formally, the invariant learning on the node level is to minimize:
\begin{equation}
\label{eq:ood_node}
  \mathop{\min}_{f}\mathop{\max}_{e \in \mathcal{E}_{all}} \mathcal{R}^{e}(f)
\end{equation}
where $\mathcal{R}^{e}(f)=\mathbb{E}_{\mathcal{G}_{v},y_{v}}^{e}[l(f(\mathcal{G}_{v}), y_{v})]$, $l(\cdot, \cdot)$ is the loss function.

Based on the above minimization objective, the trained GNN performs equally across all environments. Similarly, the goal of fairness on the graph is to have the model equally treat different demographic groups divided by the sensitive attribute $S$. In this regard, the centered node (the ego-graph) with different sensitive attribute values or under different sensitive attributes can be regarded as a graph under different environments. Naturally, a fairness problem on graphs can be formulated as a form of invariant learning.

In this work, we aim to learn fair GNNs toward various sensitive attributes in a single training session. With the formulation of invariant learning, our goal is transformed into learning GNNs invariant across different sensitive attributes and sensitive attribute values. Formally, our goal is to minimize: 
\begin{equation}
\label{eq:inv_sens}
  \mathop{\min}_{f}\mathop{\max}_{S \in \mathcal{S}_{all}} \mathcal{R}^{S}(f)
\end{equation}
where $\mathcal{R}^{S}(f)=\mathbb{E}_{\mathcal{G}_{v},y_{v}}^{S}[l(f(\mathcal{G}_{v}), y_{v})]$ is the empirical risk under sensitive attribute $S$. $\mathcal{S}_{all}$ is a set of $S$. For instance, assume that gender and race are sensitive attributes, $\mathcal{S}_{all}$ includes male, female, white, black, and yellow people environments.

\section{Present Work: FairINV}
In this section, we discuss how to learn a GNN towards fairness w.r.t. various sensitive attributes in a single training session through our proposed method FairINV. Specifically, we first give a brief overview of FairINV and then make a detailed description of the components of FairINV. Furthermore, we provide the training algorithm to shed insights into the process of FairINV.

\subsection{Overview}
FairINV focuses on the node-level task, aiming to learn GNNs invariant across various sensitive attributes within a single training session, thereby achieving fairness on graph-structured data. As shown in Figure~\ref{fig:overview}, our proposed method FairINV comprises two modules, i.e., sensitive attribute partition (SAP) and sensitive invariant learning (SIL). The SAP module partitions nodes into different subsets by inferring variant ego-subgraphs for each centered node. It should be noted that the sampling of ego-subgraphs can be disregarded due to the message-passing mechanism, which effectively aggregates the representations of neighboring nodes to update its own representation. To optimize this module, we employ the Invariant Risk Minimization (IRM) objective~\cite{arjovsky2019invariant}, maximizing it to guide the SAP module in capturing the worst-case environment. This process can be seen as inferring the sensitive attribute value of nodes. Since the maximization of the IRM objective is executed in an unsupervised manner, we can iteratively predict sensitive attributes multiple times. This iterative process enables FairINV to achieve fairness with respect to various sensitive attributes in a single training session. Due to the formulation of the fairness problem from an invariant learning perspective in Section~\ref{subsec:problem_formulate}, we can naturally tackle this problem through invariant learning. Specifically, based on the partition results of SAP, the SIL module learns a GNN invariant across different sensitive attribute partitions through a variance-based loss. The objective of being invariant across different sensitive attribute partitions implies the equitable treatment of different demographic groups, thereby achieving fair decision-making. Overall, the SAP module is akin to data augmentation, facilitating the process of the SIL module. Due to such a training paradigm, FairINV follows the same inference process as vanilla GNNs. 

\subsection{Sensitive Attributes Partition}
Existing methods are designed for the specific sensitive attribute while assuming accessible sensitive attributes. However, these methods are impractical in real-world scenarios due to legal restrictions. To overcome this challenge, there is a need to infer the sensitive attribute value for each node in an unsupervised manner. Inferring the sensitive attribute value multiple times can be regarded as obtaining multiple sensitive attribute values, e.g., gender, and race, facilitating the achievement of fairness w.r.t. various sensitive attributes. Unfortunately, learning a sensitive attribute inference model without access to the sensitive attribute ground truth is a non-trivial task. 

Inspired by the unsupervised environment inference in invariant learning~\cite{creager2021environment}, we aim to maximize variability across environments to achieve the sensitive attribute partition. Based on our formulation of the fairness problem from an invariant learning perspective, the sensitive attributes can be seen as environments in invariant learning. Nodes with different sensitive attribute values can be considered as being in different environments. Thus, maximizing variability across environments indicates inferring a worst-case sensitive attribute partition, where GNNs exhibit the worst fairness performance towards the demographic group divided by the sensitive attribute. However, directly inferring sensitive attributes partition through the aforementioned maximization objective is impractical due to the interactive nature of graph-structured data. Following the inspiration from GIL~\cite{li2022learning}, identifying variant subgraphs as auxiliary information for sensitive attribute partition may provide a desirable solution.

Following the above idea, we construct the SAP module to infer the sensitive attribute value of each node. Specifically, the SAP module consists of a pre-trained GNN backbone $\varphi$, a variant inference model $\psi$, and a sensitive attribute inference model $q$. With an expected structure identical to the GNN model to be trained, $\varphi$ serves as an Empirical Risk Minimization-trained (ERM-trained) reference model. In other words, it is trained on the node classification task in a semi-supervised manner to capture spurious correlations between variant patterns and node labels. Given an attributed graph $\mathcal{G}=(\mathcal{V}, \mathcal{E}, \textbf{X})$ with unknown sensitive attribute values, we sample an ego-graph set $\{\mathcal{G}_{v}\}_{v \in \mathcal{V}}$, where $\mathcal{G}_{v}$ is the ego-graph of the centered node $v$. $\varphi$ takes $\mathcal{G}_{v}$ as input and outputs the node representation $\textbf{h}_{v}=\varphi(\mathcal{G}_{v})$. Due to the similar process between ego-graphs sampling and the message-passing mechanism of GNN, the sampling of ego-graphs can be disregarded. For two connected nodes $u$ and $v$ in $\mathcal{G}$, $\psi$ takes the concatenation of node representations $\textbf{h}_{u}$ and $\textbf{h}_{v}$ as input to measure the variant score of edge $e_{uv}$. Assuming inferring the sensitive attribute $k$ times, the variant score in the $i$-th inferring can be formulated as: 
\begin{equation}
\label{eq:var_score}
    w_{uv}^{i} = \sigma(\psi([\textbf{h}_{u},\textbf{h}_{v}])),
\end{equation}
where $[\cdot,\cdot]$ denotes the concatenation operation, and $\sigma(\cdot)$ is a sigmoid function. In this context, $\psi$ can be implemented as a linear layer, measuring the probability that edge $e_{uv}$ belongs to the variant pattern. 

According to Eq.~\eqref{eq:var_score}, we can obtain a variant score vector $\textbf{w}^{i} \in \mathbb{R}^{|\mathcal{E}|}$, which includes variant scores for all edges. $\textbf{w}^{i}$ represents variant patterns, i.e., variant subgraphs, capturing the variant correlation between the graph structure and node labels under different sensitive attribute groups. After inferring $k$ times, we have a variant score vector set $\{\textbf{w}^{i}\}_{i=1}^{k}$, representing the variant correlation for various sensitive attributes. Accordingly, we can use these variant patterns to infer sensitive attributes. Specifically, we employ a GNN classifier as the sensitive attribute inference model $q$ to generate the sensitive attribute partition. Given the $i$-th variant score vector $\textbf{w}^{i}$, the $i$-th sensitive attribute partition  $\mathcal{P}_{i}$ can be formulated as:
\begin{equation}
\label{eq:partition}
    \mathcal{P}_{i} = q(\mathcal{G}, \textbf{w}^{i}, Y),
\end{equation}
where $\mathcal{P}_{i} \in \mathbb{R}^{|\mathcal{V}| \times t}$, and $t$ is the number of sensitive attribute groups. For instance, in the case of a sensitive attribute like gender, $t=2$.

To achieve accurate partitioning of sensitive attributes, optimizing $\psi$ and $q$ with well-defined objectives is crucial. Our goal is to capture variant patterns that result in significant performance differences across different sensitive attribute groups. Consequently, aligning with the approach of EIIL~\cite{creager2021environment}, we employ the IRM objective as the optimization objective of $\psi$ and $q$. Formally, the optimization objective of SAP can be formulated as follows:
\begin{equation}
    \label{eq:irm}
    \mathop{\max}_{\theta_{\psi}, \theta_{q}} \|\triangledown_{\overline{w}}\mathcal{R}^{S}(\overline{w}\circ \varphi,q)\|,
\end{equation}
where $\overline{w}$ denotes a constant scalar multiplier of 1 for each output dimension, and the empirical risk $\mathcal{R}^{S}(\varphi,q)$ can be formulated as:

\begin{equation}
    \mathcal{R}^{S}(\varphi,q)=\sum_{v\in \mathcal{V}}\textbf{q}_{v}(S)\mathcal{L}(\varphi(\mathcal{G}_{v}), y_{v}),
\end{equation}
where $\textbf{q}_{v}(S):q_{v}(S|\mathcal{G}_{v}, \textbf{w}^{i},y_{v})$ denotes a soft per-partition risk and is a node-level implementation of Eq.~\eqref{eq:partition}. 

Notably, the application of the IRM objective enables inferring sensitive attributes in an unsupervised manner. The inferred sensitive attributes correspond to the demographic group partition with the worst-case fairness performance. In such an unsupervised manner, we can partition sensitive attributes $k$ times to identify the top $k$ worst-case partitions, denoted by $\{\mathcal{P}_{i}\}^{k}_{i=1}$.  

\subsection{Towards Fairness via Invariant Learning}
In the SIL module, we aim to learn a fair GNN model $f$ including a GNN backbone $\Phi$ and a classifier $g$ from an invariant learning perspective. Prior works have revealed that training model $f$ in an ERM paradigm inevitably results in the capturing of spurious correlations. In our scenarios, such spurious correlations are the correlation between the sensitive attribute and node labels, being uncovered as variant patterns through the SAP module. Naturally, based on $\mathcal{P}_{i}$, we guarantee the variance across the sensitive attribute groups to optimize $f$, which is motivated by the objective of EERM~\cite{wu2022handling}. In other words, this objective guides the model to leverage the invariant patterns to yield equal performance on different sensitive attribute groups $S$. % Given an attributed graph $\mathcal{G}=(\mathcal{V}, \mathcal{E}, \textbf{X})$, we can obtain variant score vector $\{\textbf{w}_{i}\}^{k}_{i=1}$ and sensitive attributes partition $\{\mathcal{P}_{i}\}^{k}_{i=1}$ for $k$ times. In the forward of the training pipeline, $f$ takes $\mathcal{G}, \{\textbf{w}_{i}\}^{k}_{i=1}$ as input to predict node labels $\{\hat{Y}_{i}=f(\mathcal{G},\textbf{w}_{i})\}_{i=1}^{k}$.  Thus, optimization objectives of $f$ can be formulated as follows:
Given an attributed graph $\mathcal{G}=(\mathcal{V}, \mathcal{E}, \textbf{X})$, we can obtain variant score vector set $\{\textbf{w}^{i}\}_{i=1}^{k}$ and sensitive attributes partition set $\{\mathcal{P}_{i}\}_{i=1}^{k}$. In the forward of the training pipeline, $f$ takes $\mathcal{G}, \textbf{w}^{i}$ as input to predict node labels $\hat{Y}=f(\mathcal{G},\textbf{w}^{i}), i=1,2,...,k$. Thus, optimization objectives of $f$ can be formulated as follows:
\begin{equation}
    \label{eq:var_loss}
    \mathop{\min}_{\theta_{f}} Var(\{\mathcal{L}_{cls}^{S}(\hat{Y},Y)\}_{S \in \mathcal{S}_{all}}) + \alpha Mean(\{\mathcal{L}_{cls}^{S}(\hat{Y},Y)\}_{S \in \mathcal{S}_{all}}),
\end{equation}
where $Var(\cdot)$ and $Mean(\cdot)$ are variance and mean functions, respectively. The sensitive attribute group $S$ is derived from $\mathcal{P}$. $\mathcal{L}_{cls}^{S}(\cdot,\cdot)$ is the classification loss function under $S$ and we employ a binary cross-entropy function as $\mathcal{L}_{cls}$ in all experiments. $\alpha$ is a hyperparameter to balance two loss terms. 

In Eq.\eqref{eq:var_loss}, the variance loss term aims to minimize the performance difference between various sensitive attribute groups while the mean loss term ensures the predicted accuracy across all sensitive attribute groups.

\subsection{Training Algorithm}
To further help understand our proposed framework FairINV, we summarize the detailed training algorithm of FairINV, as shown in Algorithm~\ref{alg:algorithm1}.

\begin{algorithm}[t]
\caption{Training Algorithm of FairINV}
\label{alg:algorithm1}
    \begin{flushleft}
        \textbf{Input}: $\mathcal{G}=(\mathcal{V}, \mathcal{E}, \textbf{X})$ without the sensitive attribute $S$, node labels $\textbf{Y}$, the pre-trained GNN backbone $\varphi$, the variant inference model $\psi$, the sensitive attribute inference model $q$, GNN model $f$=\{$\Phi$, $g$\}, partition time $k$, and hyperparameters $\alpha$. \\
        \textbf{Output}: Trained inference GNN model $f$.
    \end{flushleft}
    \begin{algorithmic}[1] %[1] enables line numbers
    \STATE // SAP module
    \FOR{$i=1$ to $k$}
        \STATE $\textbf{H}$ $\leftarrow$ $\varphi$($\mathcal{G}$);
        \FOR{$t=1$ to $epoch_{SAP}$}
            \STATE $w_{uv}^{i}$ $\leftarrow$  $\sigma(\psi([\textbf{h}_{u},\textbf{h}_{v}]))$, $\textbf{h}_{u}, \textbf{h}_{v} \in \textbf{H}$, $e_{uv} \in \mathcal{E}$;
            \STATE // Sensitive attribute partition
            \STATE $\mathcal{P}_{i}$ $\leftarrow$ $q(\mathcal{G}, \textbf{w}^{i}, Y)$, $\textbf{w}^{i}=\{w_{uv}^{i}|u, v \in \mathcal{V}, e_{uv} \in \mathcal{E}\}$;
            \STATE Calculate loss function according to Eq.(\ref{eq:irm});
            \STATE Update parameters of $\psi$ and $q$ by gradient descent;
        \ENDFOR
        \STATE $w_{uv}^{i}$ $\leftarrow$  $\sigma(\psi([\textbf{h}_{u},\textbf{h}_{v}]))$, $\textbf{h}_{u}, \textbf{h}_{v} \in \textbf{H}$, $e_{uv} \in \mathcal{E}$;
        \STATE $\mathcal{P}_{i}$ $\leftarrow$ $q(\mathcal{G}, \textbf{w}^{i}, Y)$, $\textbf{w}^{i}=\{w_{uv}^{i}|u, v \in \mathcal{V}, e_{uv} \in \mathcal{E}\}$; 
    \ENDFOR
    \STATE Obtain $\{\mathcal{P}_{i}\}^{k}_{i=1}$, $\{\textbf{w}^{i}\}^{k}_{i=1}$;
    \STATE // SIL module
    \FOR{$t=1$ to $epoch$}
        \FOR{$i=1$ to $k$}
            \STATE $\hat{Y} \leftarrow f(\mathcal{G},\textbf{w}^{i})$;
            \STATE Calculate loss function according to Eq.\eqref{eq:var_loss} and $\mathcal{P}_{i}$;
            \STATE Accumulated loss;
        \ENDFOR
        \STATE Update parameters of $f$ by gradient descent;  
    \ENDFOR
        \STATE \textbf{return} $f$;
    \end{algorithmic}
\end{algorithm}

\section{Experiments}

\begin{table}[!t]
    \caption{Datasets statistics.}
    \centering
    \renewcommand\arraystretch{0.9}
    \resizebox{\linewidth}{!}{
    \begin{tabular}{l|ccccc}
        \toprule
        \textbf{Dataset}   & \textbf{German}    & \textbf{Bail}    & \textbf{Pokec-z}    & \textbf{Pokec-n}    & \textbf{NBA} \\
        \midrule
        \#Nodes            & 1,000              & 18,876           & 67,796              & 66,569 & 403  \\
        \#Edges            & 22,242             & 321,308          & 617,958             & 583,616 & 21,242  \\
        \#Attr.            & 27                 & 18               & 277                 & 266 & 95 \\
        Sens.              & Gender             & Race             & Region              & Region & Nationality \\
        % Labels           & Credit status      & Bail Prediction  & Future default     & Working Field       & Working Field \\
        %Training Number   & 6000               & 100              & 200                & 4000                & 3500  \\
        \bottomrule
    \end{tabular}}
    \label{tab:statistic}
\end{table}

In this section, we conduct node classification experiments on several commonly used fairness datasets, including German, Bail, Pokec-z, Pokec-n, and NBA. Table~\ref{tab:statistic} presents the statistical information of these datasets. In our experiments, we aim to answer the following three questions: \textbf{RQ1:} Can FairINV improve fairness while maintaining utility performance? \textbf{RQ2:} How does FairINV achieve fairness across various sensitive attributes in a single training session? \textbf{RQ3:} How do relevant hyperparameters and components impact FairINV?  

\subsection{Experimental Settings}
\subsubsection{Datasets}
Five real-world fairness datasets, namely German, Bail~\cite{agarwal2021towards}, Pokec-z, Pokec-n, and NBA~\cite{dai2021say}, are employed in our experiments. We give a brief overview of these datasets as follows:

\begin{itemize}
\item \textbf{German}~\cite{asuncion2007uci} is constructed by~\cite{agarwal2021towards}. Specifically, German includes clients' data in a German bank, e.g., gender, and loan amount. Nodes represent clients in the German bank. The edges in the German dataset are constructed according to individual similarity. Regarding ``gender" as the sensitive attribute, the goal of German is to classify clients into two credit risks (high or low). 

\item \textbf{Bail}~\cite{agarwal2021towards} is a defendants dataset, where defendants in this dataset are released on bail during 1990-2009 in U.S states~\cite{jordan2015effect}. We regard nodes as defendants and edges are decided by the similarity of past criminal records and demographics. Considering ``race'' as the sensitive attribute, the task is to predict whether defendants will commit a crime after release (bail vs. no bail). 

% \item \textbf{Credit}~\cite{agarwal2021towards} is a credit card user dataset~\cite{yeh2009comparisons}. where nodes represent credit card users and edges are connected based on the similarity of their spending and payment patterns. Considering ``age'' as the sensitive attribute, the task is to predict whether a user will default on their credit card payment or not (default vs. no default).

\item \textbf{Pokec-z/n}~\cite{takac2012data,dai2021say} is derived from a popular social network application in Slovakia, where Pokec-z and Pokec-n are social network data in two different provinces. Nodes denote users with features such as gender, age, interest, etc. Edge represents the friendship between users. Considering ``region'' as the sensitive attribute, the task is to predict the working field of the users.

\item \textbf{NBA}~\cite{dai2021say} is derived from a Kaggle dataset comprising approximately 400 NBA basketball players from the 2016-2017 season. Nodes denote NBA basketball players with features such as performance statistics, age, etc. Edge represents the relationship between these players on Twitter. Considering ``nationality (U.S. and overseas players)'' as the sensitive attribute, the goal is to predict whether the salary of the player is over the median.
\end{itemize}

\subsubsection{Baselines}
We compare FairINV with four state-of-the-art fairness methods, including EDITS, NIFTY, FairGNN, and FairVGNN. Among these four methods, EDITS can be summarized as the pre-processing fairness method, while the remaining baselines represent the in-processing approach. A brief overview of these methods is shown as follows:

\begin{itemize}
\item \textbf{EDITS}~\cite{dong2022edits} modify graph-structured data by minimizing the Wasserstein distance between two demographics.

\item \textbf{NIFTY}~\cite{agarwal2021towards} is a fair and stable graph representation learning method. The core idea behind NIFTY is learning GNNs to keep stable w.r.t. the sensitive attribute counterfactual.

\item \textbf{FairGNN}~\cite{dai2021say} aims to learn fair GNNs with limited sensitive attribute information. To achieve this goal, FairGNN employs the sensitive attribute estimator to predict the sensitive attribute while improving fairness via adversarial learning. 

\item \textbf{FairVGNN}~\cite{wang2022improving} learns a fair GNN by mitigating the sensitive attribute leakage using adversarial learning and weight clamping technologies.

\end{itemize}

% \usepackage{multirow}
% \usepackage[normalem]{ulem}
\begin{table*}[]
\centering
\caption{Comparison results of FairINV and baseline fairness methods on GCN backbone. In each row, the best result is indicated in \textbf{bold}, while the runner-up result is marked with an \underline{underline}. OOM: out-of-memory on a GPU with 24GB memory.}
\renewcommand\arraystretch{0.8}
\resizebox{0.8\linewidth}{!}{
\begin{tabular}{c|c|ccccc|c}
\toprule
\textbf{Datasets}                 & \textbf{Metrics} & \textbf{Vanilla GCN}  & \textbf{EDITS}       & \textbf{NIFTY}       & \textbf{FairGNN}      & \textbf{FairVGNN}     & \textbf{FairINV}      \\
\midrule
\multirow{4}{*}{\textbf{German}}  & AUC           & 65.90 $\pm$ 0.83          & {\ul 69.89 $\pm$ 3.23}   & 67.77 $\pm$ 4.30         & 67.35$\pm$2.13            & \textbf{72.38 $\pm$ 1.09} & 69.11 $\pm$ 1.80          \\
                                  & F1         & 77.32 $\pm$ 1.20          & {\ul 82.01 $\pm$ 0.91}   & 81.43 $\pm$ 0.54         & {\ul 82.01$\pm$0.26}      & 81.94 $\pm$ 0.26          & \textbf{82.36 $\pm$ 0.35} \\
                                  & $\Delta_{DP} (\downarrow)$           & 36.29 $\pm$ 4.64          & 2.38 $\pm$ 1.36          & 2.64 $\pm$ 2.25          & 3.49$\pm$2.15             & {\ul 1.44 $\pm$ 2.04}     & \textbf{0.76 $\pm$ 1.24}  \\
                                  & $\Delta_{Eo} (\downarrow)$         & 31.35 $\pm$ 4.39          & 3.03 $\pm$ 1.77          & 2.52 $\pm$ 2.88          & 3.40$\pm$2.15             & {\ul 1.51 $\pm$ 2.11}     & \textbf{0.15 $\pm$ 0.29}  \\
                                  \midrule
\multirow{4}{*}{\textbf{Bail}}    & AUC           & 87.13 $\pm$ 0.31          & {\ul 87.92 $\pm$ 1.83}   & 79.62 $\pm$ 1.80         & 87.27 $\pm$ 0.76          & 87.05 $\pm$ 0.39          & \textbf{88.53 $\pm$ 1.83} \\
                                  & F1         & 78.98 $\pm$ 0.67          & {\ul 79.45 $\pm$ 1.48}   & 67.19 $\pm$ 2.63         & 77.67 $\pm$ 1.33          & \textbf{79.56 $\pm$ 0.29} & 78.80 $\pm$ 3.71          \\
                                  & $\Delta_{DP} (\downarrow)$           & 9.18 $\pm$ 0.59           & 8.03 $\pm$ 0.97          & \textbf{3.52 $\pm$ 0.72} & 6.72 $\pm$ 0.60           & 6.31 $\pm$ 0.77           & {\ul 3.58 $\pm$ 1.61}     \\
                                  & $\Delta_{Eo} (\downarrow)$         & 4.43 $\pm$ 0.37           & 5.80 $\pm$ 0.73          & {\ul 2.82 $\pm$ 0.82}    & 4.49 $\pm$ 1.00           & 5.12 $\pm$ 1.40           & \textbf{2.15 $\pm$ 1.24}  \\
                                  \midrule
\multirow{4}{*}{\textbf{Pokec-z}} & AUC           & \textbf{76.42 $\pm$ 0.13} & \multirow{4}{*}{OOM} & 71.59 $\pm$ 0.17         & {\ul 76.02 $\pm$ 0.15}    & 75.52 $\pm$ 0.06          & 75.79 $\pm$ 0.08          \\
                                  & F1         & 70.32 $\pm$ 0.20          &                      & 67.13 $\pm$ 1.66         & 68.84 $\pm$ 3.46          & {\ul 70.45 $\pm$ 0.57}    & \textbf{70.78 $\pm$ 0.50} \\
                                  & $\Delta_{DP} (\downarrow)$            & 3.91 $\pm$ 0.35           &                      & 3.06 $\pm$ 1.85          & {\ul 2.93 $\pm$ 2.83}     & 3.30 $\pm$ 0.87           & \textbf{2.70 $\pm$ 0.96}  \\
                                  & $\Delta_{Eo} (\downarrow)$         & 4.59 $\pm$ 0.34           &                      & 3.86 $\pm$ 1.65          & \textbf{2.04 $\pm$ 2.27}  & 3.19 $\pm$ 1.00           & {\ul 2.23 $\pm$ 0.66}     \\
                                  \midrule
\multirow{4}{*}{\textbf{Pokec-n}} & AUC           & \textbf{73.87 $\pm$ 0.08} & \multirow{4}{*}{OOM} & 69.43 $\pm$ 0.31         & 73.49 $\pm$ 0.28          & 72.72 $\pm$ 0.93          & {\ul 73.55 $\pm$ 0.16}    \\
                                  & F1         & \textbf{65.55 $\pm$ 0.13} &                      & 61.55 $\pm$ 1.05         & 64.80 $\pm$ 0.89          & 62.35 $\pm$ 1.14          & {\ul 65.19 $\pm$ 0.62}    \\
                                  & $\Delta_{DP} (\downarrow)$            & 2.83 $\pm$ 0.46           &                      & 5.96 $\pm$ 1.80          & {\ul 2.26 $\pm$ 1.19}     & 4.38 $\pm$ 1.73           & \textbf{1.24 $\pm$ 0.64}  \\
                                  & $\Delta_{Eo} (\downarrow)$         & 3.66 $\pm$ 0.43           &                      & 7.75 $\pm$ 1.53          & {\ul 3.21 $\pm$ 2.28}     & 6.74 $\pm$ 1.87           & \textbf{2.80 $\pm$ 0.78}  \\
                                  \midrule
\multirow{4}{*}{\textbf{NBA}}     & AUC           & 66.09 $\pm$ 0.98          & 65.91 $\pm$ 5.19         & {\ul 68.88 $\pm$ 0.93}   & \textbf{72.53 $\pm$ 0.96} & 64.73 $\pm$ 2.34          & 66.18 $\pm$ 2.84          \\
                                  & F1         & 61.32 $\pm$ 2.53          & 61.42 $\pm$ 8.26         & {\ul 67.41 $\pm$ 2.92}   & 60.18 $\pm$ 20.18         & 60.32 $\pm$ 5.79          & \textbf{67.56 $\pm$ 1.30} \\
                                  & $\Delta_{DP} (\downarrow)$            & 28.80 $\pm$ 4.17          & 6.09 $\pm$ 5.1           & 5.41 $\pm$ 2.78          & 6.44 $\pm$ 6.74           & {\ul 3.48 $\pm$ 3.62}     & \textbf{1.98 $\pm$ 3.15}  \\
                                  & $\Delta_{Eo} (\downarrow)$         & 20.00 $\pm$ 9.43          & 6.0 $\pm$ 5.33           & 3.43 $\pm$ 1.78          & \textbf{2.64 $\pm$ 2.61}  & 3.33 $\pm$ 3.65           & {\ul 2.67 $\pm$ 3.89}    \\
                                  \midrule
\end{tabular}}
\label{tab:comparison}
\end{table*}

\subsubsection{Evaluation Metrics} 
To evaluate the utility performance, we use AUC and F1 scores. Additionally, we employ two commonly used fairness metrics, i.e., $\Delta_{DP} = \lvert P(\hat{y}=1 \lvert s=0) - P(\hat{y}=1 \lvert s=1) \rvert$~\cite{dwork2012fairness} and $\Delta_{EO} = \lvert P(\hat{y}=1 \lvert y=1,s=0) - P(\hat{y}=1 \lvert y=1,s=1) \rvert$~\cite{hardt2016equality}, to evaluate the fairness performance.

\subsubsection{Implementation Details} For all methods, including FairINV, we use a multi-layer GNN model consisting of a GNN backbone $\Phi$ and a 1-layer linear classifier $g$. To validate the generalizability of FairINV on various backbones,  we employ the following GNN backbones: a 1-layer GCN~\cite{kipf2016semi}, a 1-layer GIN~\cite{xu2018powerful}, and a 2-layer GraphSAGE~\cite{hamilton2017inductive}. Here, the hidden dimension of all GNN backbones is set to 16 for all datasets. Hyperparameter settings for all baseline methods adhere to the guidelines provided by the respective authors. We conduct all experiments 5 times and reported average results.

For FairINV, we utilize the Adam optimizer with the learning rate $lr=1 \times 10^{-2}$, epochs=1000, and the weight decay = $1 \times 10^{-5}$. Using the same optimizer, the learning rate $lr_{sp}$ for training the SAP modules are set to $\{0.1, 0.1, 0.01, 0.5, 0.1\}$ for German, Bail, Pokec-z, Pokec-n, and NBA datasets, respectively. Meanwhile, we set the balanced parameter $\alpha$ to $\{10, 10, 10, 1, 1\}$ for German, Bail, Pokec-z, Pokec-n, and NBA datasets, respectively. The partition times $k$ and the number of sensitive attribute groups $t$ are fixed at 3 and 2 for all datasets. In the SAP module, a 1-layer linear layer is used as the variant inference model $\psi$. We employ a model with the same structure as the GNN model ($\Phi$ and $g$) as a sensitive attribute inference model $q$. Meanwhile, $\varphi$ has the same structure as the GNN model ($\Phi$ and $g$) and is trained by minimizing the cross-entropy loss function. For $\varphi$ and the SAP module, we set the training epoch to 500. Due to all baselines using the sensitive attribute, FairINV incorporates the sensitive attribute into the original node features for fair comparison. Moreover, all evaluations of FairINV are conducted on a single NVIDIA RTX 4090 GPU with 24GB memory. All models are implemented with PyTorch and PyTorch-Geometric.

\subsection{Comparison Study}
To answer \textbf{RQ1}, we conduct a comparison study between FairINV and four baseline methods for the node classification task across five datasets. Specifically, we verify the effectiveness of FairINV on three GNN backbones, i.e., GIN, and GraphSAGE. Limited by the space, we only present the comparison results on the GCN backbone and provide more results in Appendix~\ref{apd:exp_gnn_backbone}. As shown in Table~\ref{tab:comparison}, the following observations can be seen: (1) FairINV outperforms all baseline methods in terms of both utility and fairness in most cases. (2) In instances where FairINV exhibits relatively lower performance, the best-performing baseline method surpasses FairINV by a slight margin. (3) FairINV improves fairness while maintaining utility performance, as evidenced by the performance improvement compared with vanilla GCN.

The first two observations verify the effectiveness of FairINV on fairness performance, simultaneously showcasing the state-of-the-art performance achieved by FairINV. As for the last observation, the potential explanation lies in FairINV's adherence to the invariance principle~\cite{rojas2018invariant,li2022learning,chen2022learning}, i.e., (1) sufficiency property and (2) invariance property. The sufficiency property emphasizes the necessity of adequate predictive abilities for the downstream task, which explains the preservation of the utility performance of FairINV. Meanwhile, the invariance property assumes consistency across different environments, signifying the invariance across the sensitive attribute groups in FairINV. Consequently, this property serves as the underlying reason for FairINV's superior fairness performance. Overall, leveraging invariant learning, FairINV captures invariant subgraphs with sufficient information for the downstream task while learning to be invariant across different sensitive attribute groups. Thus, FairINV improves fairness while preserving utility performance. In addition, as shown in Appendix~\ref{apd:exp_gnn_backbone}, similar results can be observed from the experiments on GIN and GraphSAGE backbones.

\subsection{Generalizing to Various Sensitive Attributes}
To answer \textbf{RQ2}, we generalize FairINV to various sensitive attribute scenarios. Specifically, we employ FairINV once to train a GNN model and then evaluate the fairness performance of this GNN model toward various sensitive attributes. Table~\ref{tab:var_sens} presents the results of various sensitive attributes and inferior results compared to vanilla GCN are marked with a gray background. We only present results on four datasets except for the NBA dataset due to the lack of suitable node features as the sensitive attribute. When the sensitive attribute is ``Age'', we set the median of age as the threshold to obtain binary values for the sensitive attribute. Furthermore, we provide the comparison results of FairINV and baseline methods in multi-sensitive attribute scenarios, as detailed in the Appendix~\ref{apd:exp_multi_sens}. 

We make the following observations from this table: (1) FairINV achieves superior performance compared with vanilla GCN in terms of both fairness and utility. This observation demonstrates that FairINV improves the fairness of GNNs towards various sensitive attributes in a single training session. (2) In some instances, FairINV exhibits slightly inferior fairness performance compared to vanilla GCN. We attribute this to the fact that the sensitive attribute groups partitioned by the SAP module are unrelated to the sensitive attributes we have selected. This is primarily due to the model itself making fairly equitable decisions concerning the sensitive attributes we have chosen. In other words, when grouped according to the selected sensitive attributes, the variability values are relatively small. Consequently, the sensitive attribute groups partitioned by SAP when maximizing variability are unrelated to the groups corresponding to such low variability. For the results with ``Age'' as the sensitive attribute on the Pokec-n dataset, despite decisions of the model being extremely unfair with respect to the sensitive attribute, FairINV still does not improve fairness. We attribute this to the aggressive partitioning of age into binary-sensitive attributes.
% \usepackage{multirow}
\begin{table*}[!t]
\centering
\caption{Results of various sensitive attributes. The results in which FairINV exhibits inferior performance compared to vanilla GCN are highlighted with a \colorbox{gray!30}{gray} background. Sens.Attr.: Sensitive Attribute}
\renewcommand\arraystretch{0.9}
\resizebox{0.9\linewidth}{!}{
\begin{tabular}{c|c|cc|cc|cc|cc}
\toprule
\textbf{Datasets}                  & {} & \textbf{Vanilla GCN} & \textbf{FairINV}                     & \textbf{Vanilla GCN} & \textbf{FairINV}                     & \textbf{Vanilla GCN} & \textbf{FairINV}                     & \textbf{Vanilla GCN} & \textbf{FairINV}                     \\
\midrule
                                   & Sens.Attr.          & \multicolumn{2}{c}{Age}                                     & \multicolumn{2}{c}{Gender}                                  & \multicolumn{2}{c}{ForeignWorker}                           & \multicolumn{2}{c}{Single}                                  \\
                                   & AUC                  & 65.90 ± 0.83         & 69.11 ± 1.80                         & 65.90 ± 0.83         & 69.11 ± 1.80                         & 65.90 ± 0.83         & 69.11 ± 1.80                         & 65.90 ± 0.83         & 69.11 ± 1.80                         \\
                                   & F1                   & 77.32 ± 1.20         & 82.36 ± 0.35                         & 77.32 ± 1.20         & 82.36 ± 0.35                         & 77.32 ± 1.20         & 82.36 ± 0.35                         & 77.32 ± 1.20         & 82.36 ± 0.35                         \\
                                   & $\Delta_{DP} (\downarrow)$               & 20.18 ± 5.17         & 0.48 ± 0.38                          & 36.29 ± 4.64         & 0.76 ± 1.24                          & 8.45 ± 7.05          & 2.31 ± 3.81                          & 34.13 ± 3.29         & 2.64 ± 4.15                          \\
\multirow{-5}{*}{\textbf{German}}  & $\Delta_{EO} (\downarrow)$             & 15.83 ± 3.47         & 0.46 ± 0.52                          & 31.35 ± 4.39         & 0.15 ± 0.29                          & 5.66 ± 2.61          & 2.34 ± 4.09                          & 27.26 ± 4.96         & 1.48 ± 2.96                          \\
\midrule
                                   & Sens.Attr.          & \multicolumn{2}{c}{Race}                                    & \multicolumn{2}{c}{Gender}                                  & \multicolumn{2}{c}{MARRIED}                                 & \multicolumn{2}{c}{WORKREL}                                 \\
                                   & AUC                  & 87.13 ± 0.31         & 88.53 ± 1.83                         & 87.13 ± 0.32         & 88.53 ± 1.83                         & 87.13 ± 0.33         & 88.53 ± 1.83                         & 87.13 ± 0.34         & 88.53 ± 1.83                         \\
                                   & F1                   & 78.98 ± 0.67         & \cellcolor[HTML]{D9D9D9}78.80 ± 3.71 & 78.98 ± 0.68         & \cellcolor[HTML]{D9D9D9}78.80 ± 3.71 & 78.98 ± 0.69         & \cellcolor[HTML]{D9D9D9}78.80 ± 3.71 & 78.98 ± 0.70         & \cellcolor[HTML]{D9D9D9}78.80 ± 3.71 \\
                                   & $\Delta_{DP} (\downarrow)$               & 9.18 ± 0.59          & 3.58 ± 1.61                          & 11.51 ± 0.22         & \cellcolor[HTML]{D9D9D9}12.09 ± 5.58 & 2.36 ± 0.54          & \cellcolor[HTML]{D9D9D9}3.40 ± 1.23  & 0.34 ± 0.09          & \cellcolor[HTML]{D9D9D9}0.55 ± 0.47  \\
\multirow{-5}{*}{\textbf{Bail}}    & $\Delta_{EO} (\downarrow)$             & 4.43 ± 0.37          & 2.15 ± 1.24                          & 1.95 ± 0.22          & \cellcolor[HTML]{D9D9D9}3.53 ± 2.12  & 3.13 ± 0.52          & \cellcolor[HTML]{D9D9D9}4.81 ± 2.34  & 1.25 ± 0.28          & 1.18 ± 0.67                          \\
\midrule
                                   & Sens.Attr.          & \multicolumn{2}{c}{Gender}                                  & \multicolumn{2}{c}{Region}                                  & \multicolumn{2}{c}{Age}                                     & \multicolumn{2}{c}{Hair color indicator}                    \\
                                   & AUC                  & 76.42 ± 0.13         & \cellcolor[HTML]{D9D9D9}75.79 ± 0.08 & 76.42 ± 0.13         & \cellcolor[HTML]{D9D9D9}75.79 ± 0.08 & 76.42 ± 0.13         & \cellcolor[HTML]{D9D9D9}75.79 ± 0.08 & 76.42 ± 0.13         & \cellcolor[HTML]{D9D9D9}75.79 ± 0.08 \\
                                   & F1                   & 70.32 ± 0.20         & 70.78 ± 0.50                         & 70.32 ± 0.20         & 70.78 ± 0.50                         & 70.32 ± 0.20         & 70.78 ± 0.5                          & 70.32 ± 0.20         & 70.78 ± 0.50                         \\
                                   & $\Delta_{DP} (\downarrow)$               & 3.15 ± 0.24          & 2.38 ± 1.03                          & 3.91 ± 0.35          & 2.70 ± 0.96                          & 33.09 ± 0.57         & 27.49 ± 3.29                         & 18.70 ± 0.9          & 15.63 ± 1.09                         \\
\multirow{-5}{*}{\textbf{Pokec-z}} & $\Delta_{EO} (\downarrow)$             & 5.25 ± 0.48          & 5.08 ± 0.98                          & 4.59 ± 0.34          & 2.23 ± 0.66                          & 36.32 ± 0.7          & 29.19 ± 3.68                         & 18.73 ± 0.8          & 14.31 ± 1.76                         \\
\midrule
                                   & Sens.Attr.          & \multicolumn{2}{c}{Gender}                                  & \multicolumn{2}{c}{Region}                                  & \multicolumn{2}{c}{Age}                                     & \multicolumn{2}{c}{Hair color indicator}                    \\
                                   & AUC                  & 73.87 ± 0.08         & \cellcolor[HTML]{D9D9D9}73.55 ± 0.16 & 73.87 ± 0.08         & \cellcolor[HTML]{D9D9D9}73.55 ± 0.16 & 73.87 ± 0.08         & \cellcolor[HTML]{D9D9D9}73.55 ± 0.16 & 73.87 ± 0.08         & \cellcolor[HTML]{D9D9D9}73.55 ± 0.16 \\
                                   & F1                   & 65.55 ± 0.13         & \cellcolor[HTML]{D9D9D9}65.19 ± 0.62 & 65.55 ± 0.13         & \cellcolor[HTML]{D9D9D9}65.19 ± 0.62 & 65.55 ± 0.13         & \cellcolor[HTML]{D9D9D9}65.19 ± 0.62 & 65.55 ± 0.13         & \cellcolor[HTML]{D9D9D9}65.19 ± 0.62 \\
                                   & $\Delta_{DP} (\downarrow)$               & 6.36 ± 0.20          & \cellcolor[HTML]{D9D9D9}6.44 ± 0.70  & 2.83 ± 0.46          & 1.24 ± 0.64                          & 40.05 ± 0.73         & \cellcolor[HTML]{D9D9D9}40.11 ± 0.83 & 14.96 ± 0.68         & 12.8 ± 1.90                          \\
\multirow{-5}{*}{\textbf{Pokec-n}} & $\Delta_{EO} (\downarrow)$             & 13.18 ± 0.41         & 13.13 ± 1.26                         & 3.66 ± 0.43          & 2.80 ± 0.78                          & 42.82 ± 0.60         & \cellcolor[HTML]{D9D9D9}43.53 ± 0.50 & 12.49 ± 0.51         & 12.3 ± 1.65    \\
\bottomrule
\end{tabular}}
\label{tab:var_sens}
\end{table*}

\subsection{Ablation Study}
\begin{figure}[!t]
  \centering
  \includegraphics[width=\linewidth]{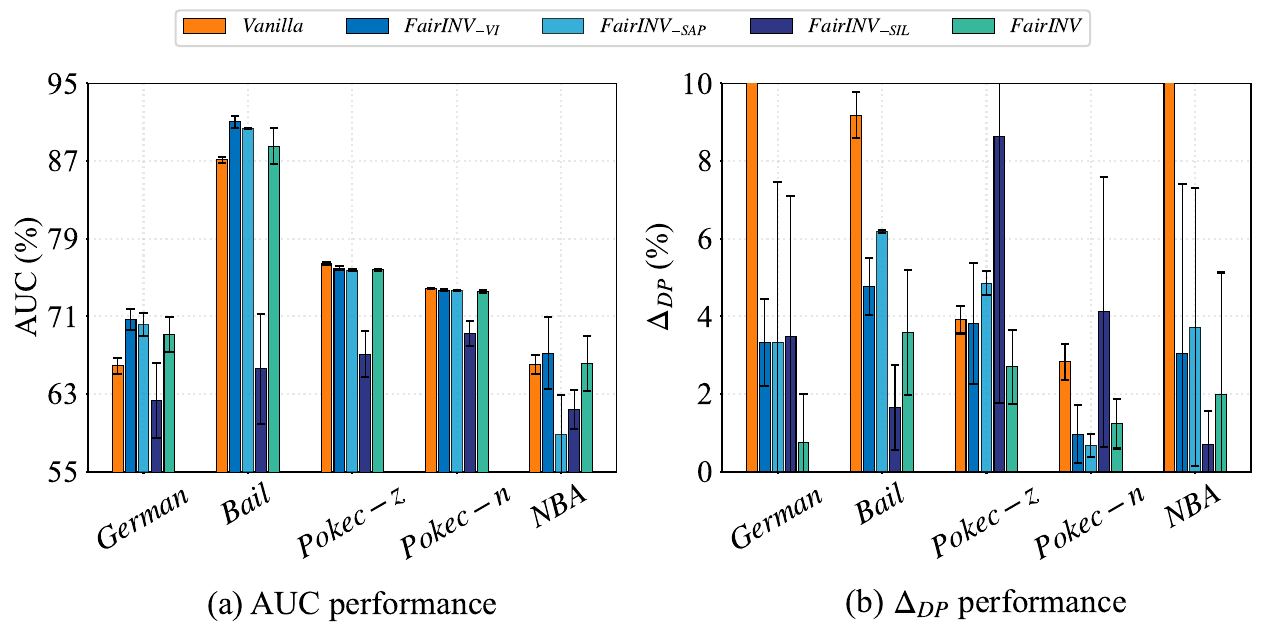}
  \caption{The results of ablation study on all datasets.}
  \label{fig:ablation}
\end{figure}

To answer \textbf{RQ3}, we conduct an ablation study to investigate the impact of each component of FairINV on improving fairness and maintaining utility. Specifically, we investigate the effect of three components including the variant inference model $\psi$, the SAP module, and the SIL module, denoted by $FairINV_{-VI}$, $FairINV_{-SAP}$, and $FairINV_{-SIL}$. $FairINV_{-VI}$ removes $\psi$, replacing $\textbf{w}$ predicted by $\psi$ with random numbers. $FairINV_{-SAP}$ removes the SAP module, replacing $\mathcal{P}$ predicted by $q$ with the sensitive attribute ground truth. $FairINV_{-SIL}$ removes the SIL module, replacing the objective shown in Eq.~\eqref{eq:var_loss} with minimizing the IRM objective shown in Eq.~\eqref{eq:irm}. Figure~\ref{fig:ablation} presents the ablation results on five datasets. 

From this figure, we observe that the removal of the SIL module leads to a decline in both utility and fairness, implying the significant impact of SIL on FairINV. Furthermore, from the results of $FairINV_{-SAP}$, even when using the ground truth of sensitive attributes to replace the predicted sensitive attribute partition $\mathcal{P}$ by SAP, the performance of FairINV is still affected. This experimental phenomenon is consistent with previous research results on invariant learning without environmental labels. Finally, we find that removing the variant inference model affects the fairness performance of FairINV, indicating the importance of the variant inference model in capturing variant patterns.

\subsection{Hyperparameters Sensitivity}
To further answer \textbf{RQ3}, we investigate the parameter sensitivity of FairINV w.r.t. two hyperparameters, i.e., the balanced parameter $\alpha$ and the learning rate $lr_{sp}$ of SAP. Notably, the setting of $lr_{sp}$ benefits from the independent training of SAP. We vary $\alpha$ and $lr_{sp}$ within the range of $\{$0.001, 0.01, 0.1, 0.5, 1, 10, 100$\}$. We only illustrate results on the Bail and Pokec-z datasets due to similar observations on other datasets. We observe that, with a wide range of variations in two parameters, the performance of FairINV remains stable. However, a sharp decline in both utility and fairness performance is noted when the value of $\alpha$ is less than 0.01.

\begin{figure}[!t]
\centering
\subfigure[AUC performance]{
\includegraphics[width=0.45\columnwidth]{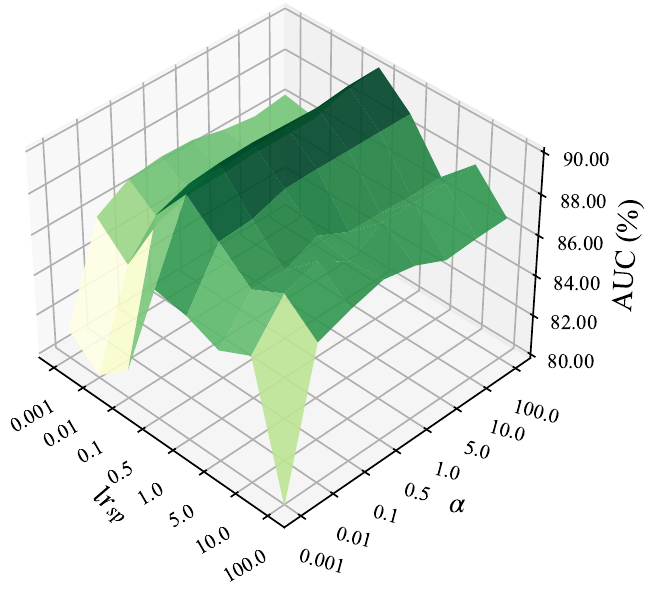}
}
% \quad
\subfigure[$\Delta_{DP}$ performance]{
\includegraphics[width=0.45\columnwidth]{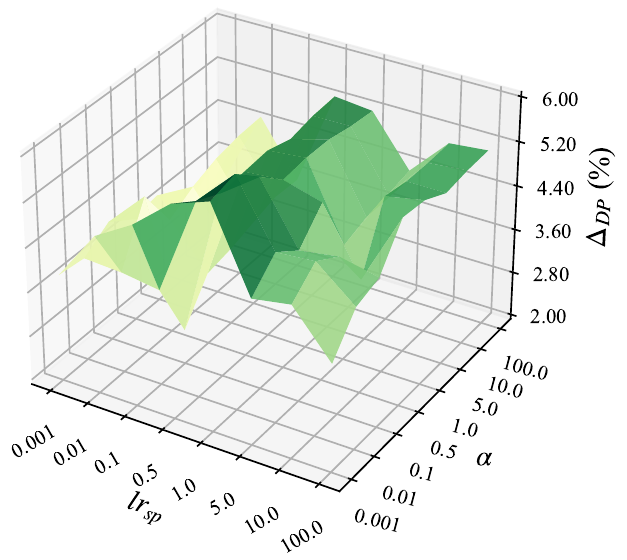}
}
\quad
\caption{Parameters sensitivity analysis on \textit{Bail}.}
\label{fig:paras_sens_bail}
\end{figure}

\begin{figure}[!t]
\centering
\subfigure[AUC performance]{
\includegraphics[width=0.45\columnwidth]{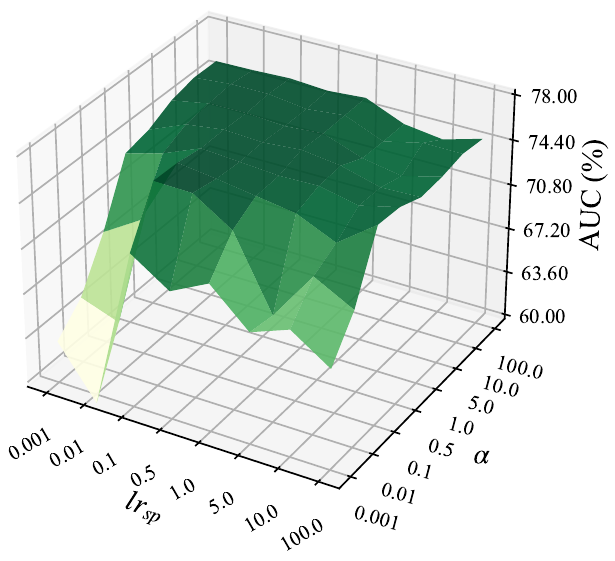}
}
% \quad
\subfigure[$\Delta_{DP}$ performance]{
\includegraphics[width=0.45\columnwidth]{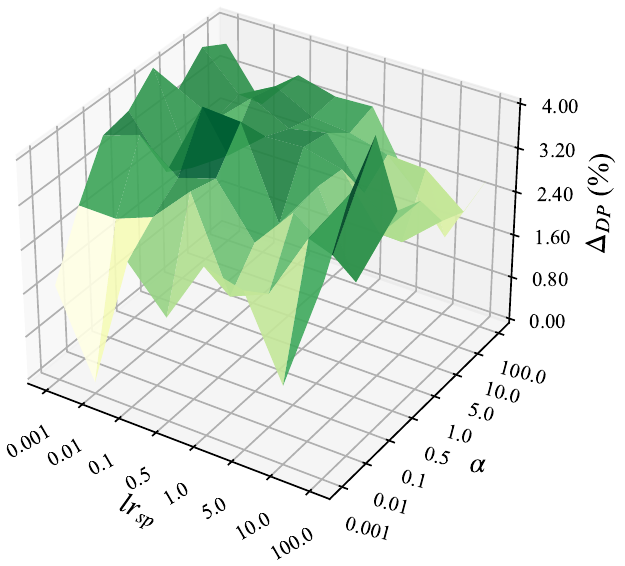}
}
\quad
\caption{Parameters sensitivity analysis on \textit{Pokec-z}.}
\label{fig:paras_sens_pokec_z}
\end{figure}

\subsection{Training Time Comparison}
To further investigate the computational cost of FairINV compared to baseline methods, we conduct a training time comparison experiment across all datasets. Specifically, we repeat each method five times and record the total training time. We set $k$ to 1 and 3 to implement two variants of FairINV, namely FairINV-1, and FairINV-3, representing FairINV trained for the single sensitive attribute and three-sensitive attribute scenarios, respectively. As shown in Table~\ref{tab:training_time}, FairINV exhibits lower computational costs on larger datasets compared to baseline methods. Although FairINV-3 requires more training time, it trains fair GNN models toward three sensitive attributes, which are unreachable for baseline methods. Furthermore, we also observe that the training time of FairINV-3 is significantly longer than that of FairINV-1. A possible explanation for this phenomenon is the high computational cost associated with the SAP module. Overall, the experimental results demonstrate that FairINV has lower computational costs than the baseline methods.

\begin{table}[!t]
\centering
\caption{Comparison of training time for both baseline methods and FairINV.}
\renewcommand\arraystretch{1.1}
\resizebox{\linewidth}{!}{
\begin{tabular}{c|cccccc}
\toprule
\textbf{Datasets} & \multicolumn{1}{c}{\textbf{EDITS}} & \multicolumn{1}{c}{\textbf{NIFTY}} & \multicolumn{1}{c}{\textbf{FairGNN}} & \multicolumn{1}{c}{\textbf{FairVGNN}} & \multicolumn{1}{c}{\textbf{FairINV-1}} & \multicolumn{1}{c}{\textbf{FairINV-3}} \\ \midrule
\textbf{German}   & 82.88 s                            & 94.91 s                            & \textbf{50.42 s}                     & 539.54 s                              & 56.20 s                                   & 114.97 s                                    \\
\textbf{Bail}     & 338.28 s                           & 106.22 s                           & 373.54 s                             & 988.19 s                              & \textbf{92.60 s}                          & 200.71 s                                    \\
\textbf{Pokec-z}  & OOM                                & 139.83 s                           & 302.11 s                             & 963.39 s                              & \textbf{113.38 s}                         & 210.29 s                                    \\
\textbf{Pokec-n}  & OOM                                & 132.03 s                           & 261.73 s                             & 1072.48 s                             & \textbf{102.33 s}                         & 191.87 s                                    \\
\textbf{NBA}      & 81.50 s                            & 92.08 s                            & \textbf{46.65 s}                     & 534.02 s                              & 57.64 s                                   & 112.75 s                                    \\ \bottomrule
\end{tabular}}
\label{tab:training_time}
\end{table}

\section{Conclusion}
In this work, we investigate the universal fairness problem, i.e., training a fair GNN toward various sensitive attributes in a single training session. To address this problem, we first formulate such a problem from a graph invariant learning point of view. Then, we propose a universal graph fairness approach, namely, FairINV. The core idea behind FairINV is to eliminate spurious correlations between the sensitive attributes and labels in a graph variant learning way. Experiments on several real-world datasets validate the effectiveness of FairINV in both fairness and utility performance. We leave validation on other downstream tasks, e.g., edge-level, as future works. In addition, due to FairINV only focusing on group fairness, future works will focus on considering fine-grained fairness, e.g., individual fairness.

\section{Acknowledgements}
The research is supported by the National Key R$\&$D Program of China under grant No. 2022YFF0902500, the Guangdong Basic and Applied Basic Research Foundation, China (No. 2023A1515011050), Shenzhen Science and Technology Program (KJZD20231023094501003), and Tencent AI Lab (RBFR2024004).

\bibliographystyle{ACM-Reference-Format}
\balance
\bibliography{sample-base}

\clearpage
\appendix
\section{Comparison for various GNN backbones}
\label{apd:exp_gnn_backbone}
To further investigate the generalizability of FairINV across various GNN backbones, we conduct comparative experiments using GIN~\cite{xu2018powerful} and GraphSAGE~\cite{hamilton2017inductive} backbones. As shown in Tables~\ref{tab:comparison_gin} and~\ref{tab:comparison_sage}, we compare FairINV with three fairness baseline methods, including NIFTY~\cite{agarwal2021towards}, FairGNN~\cite{dai2021say}, and FairVGNN~\cite{wang2022improving}. From these two tables, we can observe that FairINV consistently outperforms the three fairness baseline methods in most cases. Furthermore, upon summarizing the comparison results across the three backbones, we find that most fairness methods, including FairINV, consistently enhance both utility and fairness performance on the Bail dataset. This observation suggests an underlying relationship between fairness and utility in the Bail dataset, providing a promising avenue for future research.

\begin{table*}[!t]
\centering
\caption{Comparison results of FairINV and baseline fairness methods on GIN backbone. In each row, the best result is indicated in \textbf{bold}, while the runner-up result is marked with an \underline{underline}.}
\renewcommand\arraystretch{0.9}
\resizebox{0.8\linewidth}{!}{
\begin{tabular}{c|c|cccc|c}
\toprule
\textbf{Datasets}                 & \textbf{Metrics} & \textbf{Vanilla GIN}  & \textbf{NIFTY}       & \textbf{FairGNN}      & \textbf{FairVGNN}     & \textbf{FairINV}      \\
\midrule
\multirow{4}{*}{\textbf{German}}  & AUC           & {\ul 71.86 ± 1.55}    & 66.70 ± 4.91         & \textbf{72.78 ± 1.19} & 69.23 ± 3.08          & 70.08 ± 2.17          \\
                                  & F1         & 82.35 ± 0.55          & 80.33 ± 3.76         & 81.48 ± 1.55          & {\ul 82.41 ± 0.62}    & \textbf{82.57 ± 0.22} \\
                                  & $\Delta_{DP} (\downarrow)$           & 14.92 ± 5.52          & 5.28 ± 6.67          & 15.63 ± 5.2           & {\ul 2.71 ± 3.16}     & \textbf{1.02 ± 1.17}  \\
                                  & $\Delta_{EO} (\downarrow)$         & 8.24 ± 6.31           & 7.39 ± 8.49          & 10.0 ± 5.51           & {\ul 0.91 ± 1.41}     & \textbf{0.17 ± 0.34}  \\
                                  \midrule
\multirow{4}{*}{\textbf{Bail}}    & AUC           & 75.69 ± 7.75          & 79.49 ± 6.65         & 83.96 ± 0.61          & \textbf{86.33 ± 1.05} & {\ul 86.05 ± 0.81}    \\
                                  & F1         & 64.26 ± 8.73          & 65.20 ± 11.22        & 73.10 ± 1.28          & \textbf{87.47 ± 0.50} & {\ul 75.66 ± 3.01}    \\
                                  & $\Delta_{DP} (\downarrow)$           & 8.44 ± 2.94           & \textbf{5.38 ± 1.16} & 8.93 ± 1.63           & {\ul 6.95 ± 0.41}     & 7.35 ± 1.71           \\
                                  & $\Delta_{EO} (\downarrow)$         & 6.57 ± 1.36           & \textbf{4.00 ± 2.21} & 6.65 ± 1.77           & 6.97 ± 1.18           & {\ul 4.80 ± 1.21}     \\
                                  \midrule
\multirow{4}{*}{\textbf{Pokec-z}} & AUC           & \textbf{75.04 ± 0.39} & 72.52 ± 2.66         & 74.70 ± 1.21          & 74.51 ± 0.12          & {\ul 74.90 ± 1.28}    \\
                                  & F1         & {\ul 68.45 ± 1.23}    & 67.93 ± 1.26         & 67.30 ± 0.77          & \textbf{69.70 ± 0.57} & 67.47 ± 1.98          \\
                                  & $\Delta_{DP} (\downarrow)$           & 3.24 ± 2.09           & 3.56 ± 2.95          & 3.96 ± 1.47           & {\ul 1.93 ± 1.23}     & \textbf{1.66 ± 1.16}  \\
                                  & $\Delta_{EO} (\downarrow)$         & 4.26 ± 2.27           & 3.51 ± 2.20          & 5.22 ± 1.51           & {\ul 2.71 ± 1.20}     & \textbf{2.06 ± 0.89}  \\
                                  \midrule
\multirow{4}{*}{\textbf{Pokec-n}} & AUC           & {\ul 74.06 ± 0.62}    & 72.12 ± 1.65         & 73.25 ± 1.04          & 72.71 ± 0.48          & \textbf{74.39 ± 0.48} \\
                                  & F1         & {\ul 62.39 ± 0.51}    & 60.25 ± 4.53         & 60.88 ± 2.99          & \textbf{65.56 ± 1.03} & 62.09 ± 2.37          \\
                                  & $\Delta_{DP} (\downarrow)$           & 2.64 ± 1.28           & 3.34 ± 1.78          & {\ul 2.25 ± 1.33}     & 6.13 ± 1.59           & \textbf{1.37 ± 0.91}  \\
                                  & $\Delta_{EO} (\downarrow)$         & 6.77 ± 2.36           & 6.88 ± 2.11          & {\ul 2.68 ± 1.59}     & 7.00 ± 1.80           & \textbf{2.03 ± 2.04} \\
                                  \bottomrule
\end{tabular}}
\label{tab:comparison_gin}
\end{table*}

\begin{table*}[!t]
\centering
\caption{Comparison results of FairINV and baseline fairness methods on GraphSAGE backbone. In each row, the best result is indicated in \textbf{bold}, while the runner-up result is marked with an \underline{underline}.}
\renewcommand\arraystretch{0.9}
\resizebox{0.85\linewidth}{!}{
\begin{tabular}{c|c|cccc|c}
\toprule
\textbf{Datasets}                 & \textbf{Metrics} & \multicolumn{1}{c}{\textbf{Vanilla GraphSAGE}} & \textbf{NIFTY}       & \textbf{FairGNN}     & \textbf{FairVGNN}           & \textbf{FairINV}                          \\
\midrule
\multirow{4}{*}{\textbf{German}}  & AUC           & {\ul 74.41 ± 0.80}                        & 68.45 ± 3.8          & \textbf{75.25 ± 0.9} & 73.79 ± 1.67                & \multicolumn{1}{l}{73.64 ± 2.88}          \\
                                  & F1         & 80.74 ± 1.81                             & 77.35 ± 0.69         & 79.45 ± 2.69         & {\ul 82.20 ± 0.48}          & \multicolumn{1}{l}{\textbf{82.49 ± 0.23}} \\
                                  & $\Delta_{DP} (\downarrow)$           & 26.89 ± 6.23                             & 5.93 ± 7.03          & 27.45 ± 4.59         & {\ul 2.98 ± 2.75}           & \multicolumn{1}{l}{\textbf{0.34 ± 0.58}}  \\
                                  & $\Delta_{EO} (\downarrow)$         & 18.36 ± 6.91                             & 5.27 ± 4.02          & 20.21 ± 4.48         & {\ul 1.38 ± 0.89}           & \multicolumn{1}{l}{\textbf{0.17 ± 0.34}}  \\
                                  \midrule
\multirow{4}{*}{\textbf{Bail}}    & AUC           & 90.79 ± 1.14                             & 91.18 ± 1.32         & 91.48 ± 0.28         & \textbf{92.01 ± 0.68}       & {\ul 91.85 ± 0.43}                        \\
                                  & F1         & 80.82 ± 1.81                             & 80.54 ± 1.52         & 81.41 ± 0.54         & \textbf{83.85 ± 1.15}       & {\ul 81.59 ± 1.66}                        \\
                                  & $\Delta_{DP} (\downarrow)$           & 2.45 ± 1.31                              & \textbf{6.19 ± 1.64} & {\ul 1.52 ± 0.85}    & {\ul 3.00 ± 1.55}           & \textbf{0.49 ± 0.43}                      \\
                                  & $\Delta_{EO} (\downarrow)$         & 1.77 ± 0.68                              & \textbf{4.75 ± 1.62} & 1.44 ± 0.84          & {\ul 1.48 ± 1.34}           & {\ul \textbf{0.66 ± 0.49}}                \\
                                  \midrule
\multirow{4}{*}{\textbf{Pokec-z}} & AUC           & \textbf{78.69 ± 0.44}                    & 77.05 ± 0.53         & 77.86 ± 0.93         & {\ul 78.67 ± 0.57}          & {\ul 78.22 ± 0.68}                        \\
                                  & F1         & {\ul 70.54 ± 1.26}                       & 65.19 ± 3.18         & 68.83 ± 3.88         & \textbf{72.78 ± 0.73}       & 70.48 ± 2.32                              \\
                                  & $\Delta_{DP} (\downarrow)$           & 4.99 ± 1.41                              & {\ul 3.65 ± 0.94}    & 5.48 ± 1.15          & {\ul \textbf{3.08 ± 1.73}}  & \textbf{3.76 ± 1.29}                      \\
                                  & $\Delta_{EO} (\downarrow)$         & 5.17 ± 1.68                              & 3.87 ± 1.21          & 5.61 ± 1.48          & {\ul 3.85 ± 1.90}           & \textbf{3.46 ± 0.99}                      \\
                                  \midrule
\multirow{4}{*}{\textbf{Pokec-n}} & AUC           & {\ul 75.99 ± 0.39}                       & 72.31 ± 1.67         & 75.12 ± 1.03         & 75.22 ± 0.63                & \textbf{76.08 ± 0.31}                     \\
                                  & F1         & {\ul 63.03 ± 1.49}                       & 61.73 ± 1.4          & 64.84 ± 1.86         & {\ul \textbf{65.91 ± 1.43}} & \textbf{66.22 ± 1.61}                     \\
                                  & $\Delta_{DP} (\downarrow)$           & {\ul 1.02 ± 0.67}                        & 6.66 ± 1.40           & {\ul 1.93 ± 1.14}    & 4.94 ± 1.91                 & \textbf{0.99 ± 0.90}                       \\
                                  & $\Delta_{EO} (\downarrow)$         & 2.65 ± 1.20                               & 9.16 ± 1.86          & {\ul 2.6 ± 2.19}     & 6.26 ± 2.71                 & \textbf{1.49 ± 1.11}  \\
                                  \bottomrule
\end{tabular}}
\label{tab:comparison_sage}
\end{table*}

\section{Comparison for Multi-sensitive attributes}
\label{apd:exp_multi_sens}
We further present a comparison of FairINV and two baseline methods in various sensitive attribute scenarios, as shown in Table~\ref{tab:comparison_var_sens}. Due to the single sensitive attribute setting of these two methods, it is necessary to extend them by modifying the optimization objectives. For NIFTY~\cite{agarwal2021towards}, we simultaneously flap various sensitive attributes to construct the counterfactual graph. For FairGNN~\cite{dai2021say}, we train multiple sensitive attribute estimators and discriminators simultaneously. Although existing methods can be extended to multi-sensitive attribute scenarios, their performance might be negatively affected since they are not explicitly designed for multiple sensitive attributes. From Table~\ref{tab:comparison_var_sens}, we can observe that FairINV is the only method that can achieve fairness and maintain utility. In most cases, FairINV's fairness performance is better than baseline methods.

\begin{table*}[!t]
\centering
\caption{Comparison of FairINV and baseline methods in various sensitive attributes scenarios. Sens.Attr.: Sensitive Attribute.}
\renewcommand\arraystretch{1}
\resizebox{\linewidth}{!}{
\begin{tabular}{c|c|cccc|cccc|cccc}
\toprule
\textbf{Datesets}              & {}         & \textbf{Vanilla GCN} & \textbf{NIFTY} & \textbf{FairGNN} & \textbf{FairINV} & \textbf{Vanilla GCN} & \textbf{NIFTY} & \textbf{FairGNN} & \textbf{FairINV} & \textbf{Vanilla GCN} & \textbf{NIFTY} & \textbf{FairGNN} & \textbf{FairINV} \\
\midrule
{}     & \textbf{Sens.Attr.} & \multicolumn{4}{c}{Age}                                                     & \multicolumn{4}{c}{Gender}                                                  & \multicolumn{4}{c}{Single}                                                  \\
\multirow{4}{*}{\textbf{German}}  & \textbf{AUC}              & 65.90 ± 0.83         & 49.26 ± 6.68   & 75.13 ± 0.84     & 69.11 ± 1.80     & 65.90 ± 0.83         & 55.87 ± 8.19   & 75.69 ± 0.60     & 69.11 ± 1.80     & 65.90 ± 0.83         & 56.08 ± 8.46   & 75.69 ± 0.60      & 69.11 ± 1.80     \\
                                  & \textbf{F1}            & 77.32 ± 1.20         & 82.0 ± 0.71    & 76.83 ± 3.09     & 82.36 ± 0.35     & 77.32 ± 1.20         & 81.88 ± 0.39   & 77.91 ± 5.16     & 82.36 ± 0.35     & 77.32 ± 1.20         & 82.03 ± 0.37   & 78.17 ± 5.15     & 82.36 ± 0.35     \\
                                  & \textbf{$\Delta_{DP} (\downarrow)$}              & 20.18 ± 5.17         & 0.13 ± 0.25    & 20.67 ± 3.40      & 0.48 ± 0.38      & 36.29 ± 4.64         & 1.75 ± 1.82    & 36.07 ± 5.75     & 0.76 ± 1.24      & 34.13 ± 3.29         & 1.19 ± 1.71    & 33.37 ± 7.83     & 2.64 ± 4.15      \\
                                  & \textbf{$\Delta_{EO} (\downarrow)$}            & 15.83 ± 3.47         & 0.34 ± 0.67    & 18.24 ± 3.95     & 0.46 ± 0.52      & 31.35 ± 4.39         & 1.53 ± 0.93    & 27.84 ± 5.71     & 0.15 ± 0.29      & 27.26 ± 4.96         & 0.61 ± 0.69    & 25.91 ± 7.93     & 1.48 ± 2.96      \\
                                  \midrule
{}     & \textbf{Sens.Attr.} & \multicolumn{4}{c}{Hair color indicator}                                  & \multicolumn{4}{c}{Region}                                                  & \multicolumn{4}{c}{AGE}                                                     \\
\multirow{4}{*}{\textbf{Pokec-z}} & \textbf{AUC}              & 76.42 ± 0.13         & 74.04 ± 0.46   & 70.39 ± 1.07     & 75.79 ± 0.08     & 76.42 ± 0.13         & 74.04 ± 0.46   & 70.25 ± 0.89     & 75.79 ± 0.08     & 76.42 ± 0.13         & 74.04 ± 0.46   & 70.31 ± 1.07     & 75.79 ± 0.08     \\
                                  & \textbf{F1}            & 70.32 ± 0.20         & 69.90 ± 0.34    & 43.73 ± 24.28    & 70.78 ± 0.50      & 70.32 ± 0.20         & 69.9 ± 0.34    & 41.98 ± 23.68    & 70.78 ± 0.5      & 70.32 ± 0.20         & 69.90 ± 0.34    & 41.34 ± 28.04    & 70.78 ± 0.50      \\
                                  & \textbf{$\Delta_{DP} (\downarrow)$}              & 18.7 ± 0.90           & 17.29 ± 7.62   & 10.80 ± 7.58      & 15.63 ± 1.09     & 3.91 ± 0.35          & 8.15 ± 3.28    & 2.80 ± 1.11       & 2.70 ± 0.96       & 33.09 ± 0.57         & 35.71 ± 13.32  & 22.08 ± 18.8     & 27.49 ± 3.29     \\
                                  & \textbf{$\Delta_{EO} (\downarrow)$}            & 18.73 ± 0.80          & 14.81 ± 6.47   & 10.19 ± 7.21     & 14.31 ± 1.76     & 4.59 ± 0.34          & 7.57 ± 2.93    & 3.45 ± 1.38      & 2.23 ± 0.66      & 36.32 ± 0.70          & 33.68 ± 13.4   & 23.42 ± 19.31    & 29.19 ± 3.68   \\
                                  \bottomrule
\end{tabular}}
\label{tab:comparison_var_sens}
\end{table*}

\end{document}